\documentclass[twocolumn]{svjour3}          
\def\makeheadbox{\relax}
\usepackage{mathptmx}
\usepackage{amsmath}
\usepackage{amssymb}
\usepackage{pdflscape}
\usepackage{booktabs,makecell,multirow}
\usepackage{enumitem} 
\usepackage{tikz}
\usepackage[english]{babel}
\usetikzlibrary{arrows,positioning} 
\usetikzlibrary{shadows.blur}
\tikzset{
    >=stealth',
    box/.style={
           rectangle,
           fill=white,
           draw=black, thick,
           text width=7.5em,
           minimum height=2em,
           blur shadow={shadow blur steps=5,shadow blur extra rounding=1.3pt},
           text centered},
    titlebox/.style={
           rectangle,
           fill=white,
           draw=gray,
           minimum height=2em,
           text centered},
    nobox/.style={
           rectangle,
           fill=white,
           draw=gray,
           text width=7.5em,
           minimum height=2em,
           text centered},
    dummy/.style={
           minimum height=2em},
    dummyedge/.style={
           shorten <=-1em
    },
    dummyhor/.style={
           shorten <=-.4em
    },
    pil/.style={
           <-,
           thick,
           shorten <=2pt,
           shorten >=2pt,},
    pil2/.style={
           <-,
           dashed,
           thick,
           shorten <=2pt,
           shorten >=2pt,}
}
\usepackage{xcolor} 
\usepackage[utf8]{inputenc}
\usepackage{hyperref}
\usepackage{graphicx}
\begin{document}

\newcommand{\revised}[1]{#1}

\title{A snapshot on nonstandard supervised learning problems}
\subtitle{Taxonomy, relationships and methods}


\author{David Charte
  \and Francisco Charte
  \and Salvador Garc{\'i}a
  \and Francisco Herrera
}


\institute{D. Charte \at
              Universidad de Granada, Granada, Spain \\             
              \email{fdavidcl@ugr.es}          
           \and
           F. Charte \at
           Universidad de Ja{\'e}n, Ja{\'e}n, Spain \\
           \email{fcharte@ujaen.es}
           \and
           S. Garc{\'i}a \at
              Universidad de Granada, Granada, Spain \\
           \email{salvagl@decsai.ugr.es}
           \and
           F. Herrera \at
              Universidad de Granada, Granada, Spain \\
           \email{herrera@decsai.ugr.es}
}


\date{This is a pre-print of an article published in Progress in Artificial Intelligence. The final authenticated version is available online at: \url{https://doi.org/10.1007/s13748-018-00167-7}}

\maketitle

\begin{abstract}
Machine learning is a field which studies how machines can alter and adapt their behavior, improving their actions according to the information they are given. This field is subdivided into multiple areas, among which the best known are supervised learning (e.g. classification and regression) and unsupervised learning (e.g. clustering and association rules).

Within supervised learning, most studies and research are focused on well known standard tasks, such as binary classification, multiclass classification and regression with one dependent variable. However, there are many other less known problems. These are what we generically call nonstandard supervised learning problems. The literature about them is much more sparse, and each study is directed to a specific task. Therefore, the definitions, relations and applications of this kind of learners are hard to find.

The goal of this paper is to provide the reader with a broad view on the distinct variations of nonstandard supervised problems. A comprehensive taxonomy summarizing their traits is proposed. A review of the common approaches followed to accomplish them and their main applications is provided as well.

\keywords{Machine learning \and Supervised learning \and Nonstandard learning}
\subclass{MSC 68T05 \and MSC 68T10}
\end{abstract}

\section{Introduction}
\label{intro}

According to Mitchell \cite{learning-mitchell}, a machine is said to learn from experience $E$ related to a class of tasks $T$ and performance metric $P$, when its performance at tasks in $T$  improves according to $P$ after experience $E$.

Supervised learning is one of the fundamental areas of machine learning \cite{learning-marsland}. From object detection to ecological modeling to emotion recognition, it covers all kinds of applications. It essentially consists in learning a function by training with a set of input-output pairs. The training stage can be seen as $E$ in the previous definition, and the specific task $T$ may vary, but usually involves predicting an appropriate output given a new input.

Traditionally, supervised learning problems have been spread into two categories: classification and regression \cite{classification,pattern-rec}. In the first, information is divided into discrete categories, while the latter involves patterns associated to a value in a continuous spectrum.

These problems can be processed by learning from a training dataset, which is composed of instances. Typically, these instances or samples take the form $(x, y)$ where $x$ is a vector of values in the space of input variables and $y$ is a value in the target variable. Each problem can be described by the type of its instances: inputs will usually belong to a subset of $\mathbb R^n$, and outputs will take values in a specific one-dimensional set, finite or continuous. Once trained, the obtained model can be used to predict the target variable on unseen instances.

Standard classification problems are those where labels are either binary or multiclass \cite{classification-duda,multiclass}. In the binary case, an instance can only be associated with one of two values: positive or negative, which is equivalent to 0 or 1. For example, email messages may be classified into spam or legit, and tumours can be categorized as either benign or malign. Multiclass problems, on the other hand, involve any finite number of classes. That is, any given instance will belong to one of possibly many categories, which is equivalent to it being assigned a natural number below a convenient threshold. As an example, a photograph of a plant or a sound recording from an animal could correspond to one of a variety of species. 

A standard regression problem \cite{learning-tibshirani,regression} consists in finding a function which is able to predict, for a given example, a real value among a continuous range, usually an interval or the set of real numbers $\mathbb R$. For example, the height of a person may be estimated out of several characteristics such as age or country of origin.

Even though these standard problems are applicable in a multitude of cases, there are situations whose correct modeling requires modifications of their structure. For example, a newspaper article can be categorized according to its contents, but it could be desirable to assign several categories simultaneously. Similarly, a social media post could be described by not one but two input vectors, an image and a piece of text. These special circumstances cannot be covered by the traditional one-vector input and one-dimensional output schema. As a consequence, since performance metrics which measure improvements in standard tasks assume the common structure, they lose applicability or sense in these cases. Thus, not only new techniques are needed to tackle the problems, but also new ways of measuring and comparing their success.

This work studies variations on classic supervised problems where the traditional structure is not obeyed, which we call nonstandard variations. These emerge when the structure of the classical components of the problems does not suffice to describe complex situations, such as multiplicity of inputs or outputs, or order restrictions. As a consequence, this manuscript does not cover other singular supervised problems, such as high dimensionality of the feature space \cite{highdim} or unbalanced training sets \cite{imbalanced,imbalanced-krawczyk}, nor time-dependent problems, such as data streams \cite{streams,streams2} or time series \cite{timeseries}.

The rest of the paper is structured as follows. Section \ref{sec:definitions} formally defines and describes each nonstandard variation. This is followed by Section \ref{sec:taxonomy} establishing relations among the introduced problems and proposing a taxonomy of them. Section \ref{sec:algorithms} describes the most common techniques used to solve them. After that, Section \ref{sec:applications} enumerates popular applications of each problem. \revised{Section \ref{sec:othervariations} covers other variations further from the ones previously detailed.} Lastly, Section \ref{sec:conclusions} draws some conclusions.

\section{Definitions of nonstandard variations}
\label{sec:definitions}

The problems introduced in this section are generalizations over the traditional versions of classification and regression. The focus is on fully supervised problems, where inputs are always paired with outputs during training. An alternative taxonomy based on different supervision models is introduced in \cite{weak-nonstandard}.

\subsection{Notation}
\label{sec:notation}

In this work we will establish a notation which intends to be as simple to understand as possible, while being able to encompass every nonstandard variation. First, any supervised learning problem consists in finding a function which will classify, rank or perform regression. It will be noted as
\begin{equation}
  f: X\rightarrow Y
\end{equation}
where $X$ is an input set, or domain, and $Y$ is an output set, or codomain. It will be assumed that a training dataset $S$ is provided, including a finite number of input-output pairs:
\begin{equation}
  (x, y)\in S\subset X\times Y~.
\end{equation}
This way, a learning algorithm will be able to generate the desired function $f$. An additional notation will be the set of labels $\mathcal L$ where convenient. 

For example, in standard binary classification $X\subset\mathbb R^n$ and $Y=\mathcal L=\{0, 1\}$. Similarly, standard regression problems can be defined with the same kind of $X$ set and $Y\subset \mathbb R$. Thus, we can define very distinct supervised problems by particularizing sets $X$ or $Y$ in different ways.

Other usual notations are based in probability theory, thus involving random variables and probability distributions \cite{gaussianproc,learning-murphy}. In that case, $X$ and $Y$ would be the sample spaces of the input and output variables $\mathbf X$ and $\mathbf Y$, respectively. Predictors would usually attempt to infer a discriminant model $P(\mathbf Y|\mathbf X)$ from the training dataset.

\subsection{Multi-instance}
\label{sec:minstance}

The multi-instance (MI) framework \cite{mil} assumes a single feature space for all instances, but each training pattern may consist of more than one instance. In this case, a training pattern is composed of a finite multiset or \emph{bag} of instances and a label. Formally, assuming instances are drawn from a set $A\subset\mathbb R^n$, the domain can be described as follows:
\begin{equation}
  X=\left\{b\subset A\mid b \mbox{ finite}\right\}~.
  \end{equation} 

In this case, the learning algorithm will not know labels associated to each instance but to a bag of them. In addition to this, not all instances may share the same relevance or are equally related to the label.

Some MI problems assume that hidden labels are present for each instance in a bag: for example, a training set of drug tests where, for each test, several drug types are analyzed. Additionally, a typical MI assumption in the binary scenario states that a bag is positive when at least one of its instances is positive, and it is negative otherwise \cite{mil-assumptions}.

Other MI problems differ in that a per-instance labeling may not be possible or may not make sense: for example, if each bag represents an image and instances are image segments, class \emph{beach} can only apply to bags with water and sand segments, but it cannot apply to an individual instance.

\subsection{Multi-view}
\label{sec:mview}

A learning problem is considered to be multi-view (MV) \cite{mviewl} when inputs are composed of several components of very different nature. 

For example, if a learning pattern consists of an image as well as a piece of text representing the same instance, they can be seen as two \emph{views} on it. In that case, images and texts would belong to distinct feature spaces $A$ and $B$ respectively, an input pattern being $(a,b)\in A\times B$ . More generally, we can describe the input space as:
\begin{equation}
  X=\prod_{i=1}^{t}A_{i}~\mbox{, where }A_i\subset\mathbb R^{n_i},
  \end{equation}
where $t$ is the number of views offered by the problem and $n_i$ is the dimension of the feature space of the $i$-th view.

\subsection{Multi-label}
\label{sec:mlabel}

The multi-label (ML) learning field \cite{mlc,mltutorial} studies problems related to simultaneously assigning multiple labels to a single instance. That is, if $\mathcal L = \{l_1,\dots,l_p\}$ the codomain consists of all possible selections of these $p$ labels, also known as \emph{labelsets}:
\begin{equation}
  Y=2^{\mathcal L}\cong\left\{0,1\right\}^p~.
  \end{equation} 
As shown by this formulation, it is equivalent to think of a selection of labels as a subset of $\mathcal L$ and as a binary vector. For example, the labelset composed of the first and third labels can be represented either by $\{l_1,l_3\}$ or $(1,0,1,0,\dots,0)$.

The difference that arises when comparing ML problems to binary or multiclass ones is that labels may interact with each other. For example, a news piece classified in \emph{economy} is more likely to be labeled \emph{politics} than \emph{sports}. Similarly, a photograph labeled \emph{ocean} is less likely to have the \emph{mountains} label rather than \emph{beach}. Methods may take advantage of label co-ocurrence \cite{scumble} in order to reduce the search space when predicting a labelset. 


\subsection{Multi-dimensional}
\label{sec:mdim}

Multi-dimensional (MD) learning \cite{mdc} is a generalized classification problem where categorization is performed simultaneously along several dimensions. Each instance can belong to one of many classes in each dimension, thus the output space can be formally described as:
\begin{equation}
  Y=\mathcal L_1\times\mathcal L_2\times\dots\times\mathcal L_p,
  \end{equation}
where $\mathcal L_i$ is the label space for the $i$-th dimension. 

As with ML learning, label dimensions may be related in some way and treating them independently would only be a naive solution to the problem.

\subsection{Label distribution learning}
\label{sec:ldl}

In label distribution learning (LDL) problems \cite{ldl}, otherwise known as probabilistic class label problems \cite{ldl-prob}, any instance can be described in different degrees by each label. This can be modeled as a discrete distribution over the labels, where the probability of a label given a specific instance is called its \emph{degree of description}. Analitically, the objective is, for each instance, to predict a real-valued vector which sums exactly 1:
\begin{equation}
  Y=\left\{y\in\left[0,1\right]^p:\sum_{i=1}^p y_i = 1\right\}~.
  \end{equation}
  
In this case, we would say that the $i$-th label in $\mathcal L$ describes an instance $(x, y)$ with degree $y_i$.

\subsection{Label ranking}

In a label ranking (LR) problem \cite{lrankpairwise,lranksurvey} the objective is not to find a function able to choose one or several labels from the label space. Instead, it must evaluate their relevance for each unseen instance. The most general version of the problem involves a training set where $Y$ is the set of all partial orders of $\mathcal L$, and the obtained function also maps individual instances to partial orders. This way, for each test instance the function will output a sequence of preferences where some labels will be seen as more relevant than others. 

However, the typical situation in label ranking problems is that the orders are total, which means any two labels can always be compared. This is called a \emph{ranking} and does not exclude the possibility of ties. When ties are not allowed it is said to be a \emph{sorting} or \emph{permutation}, and can be formulated as follows:
\begin{equation}
Y=\left\{\sigma:\{1,\dots,p\}\rightarrow \mathcal L\mid\sigma\mbox{ is bijective}\right\}~,
\end{equation}
where $p$ is the amount of labels. $Y$ can also be seen as the set of all permutations of the labels in $\mathcal L$, usually known as the symmetric group of order $p$, and noted as $S_p$. 

\subsection{Multi-target regression}
\label{sec:mtarget}

A regression problem where the output space has more than just one dimension is usually called multi-target regression (MTR) and is also known as multi-output, multi-variate or multi-response \cite{moutr}. In this case, a formal description is simply that the codomain is a continuous multi-dimensional real set:
\begin{equation}
  Y=\prod_{i=1}^p Y_i~\mbox{, where }Y_i\subset\mathbb R~\forall i
  \end{equation}
and $p$ is the number of target variables.

As with other multiple target extensions, the key difference with single-target regression in this case is the possible interactions among output variables.

  
\subsection{Ordinal regression}

A problem where the target space is discrete but ordered is called ordinal regression (OR) or, alternatively, ordinal classification \cite{ord-survey}. It can be located midway between classification and regression. More specifically, it consists in labeling instances with a finite number of choices where these are ordered
\begin{equation}
Y=\left\{1,2,\dots,c\right\},~1<2<\dots<c~.
\end{equation}

In OR, the training phase consists in learning from a set of feature vectors which have a specific label associated to them, and testing can be performed over individual instances. This means that, although labels are ordered, the main objective is not to rank or sort instances as in learning to rank \cite{ltr}, but to simply classify them. The labels themselves do not provide any metric information either, they only carry qualitative information about the order among themselves.

\subsection{Monotonicity constraints}

Order relations can exist not only in the label space but in the feature space as well. Partial orders among real-valued feature vectors are always possible, and there may be cases where the order among instances is determined by just one or a few of their attributes.

When inputs as well as outputs are at least partially ordered, it is common to look for predictions which respect their order relations. In that case, the objective is to obtain a classifier or regression function which enforces the following constraint:
\begin{equation}
x_1<x_2\Rightarrow f(x_1)<f(x_2)~\forall x_1,x_2\in X~.
\end{equation}

When $Y$ is discrete the problem is usually called \emph{monotone classification} (MC), monotonic classification or ordinal classification with monotonicity constraints \cite{mc-salva}. If, on the contrary, $Y$ is continuous, it is known as \emph{isotonic regression} (IR) \cite{ir-book}.

\subsection{Absence or partiality of information}

Some problems do not directly alter the structure of $X$ and $Y$ from the standard supervised problem. Instead, they restrict which data can belong to a training set, or remove labelings from training examples. In this case, training information is presented partially or with some exclusions.

According to which kind of information is missing from the training set, a learning task  can usually be categorized as semi-supervised \cite{semi-sup}, one-class learning \cite{oneclass}, PU-learning \cite{pu-learn}, zero-shot learning \cite{zeroshot} or one-shot learning \cite{oneshot}. These are described further in Section \ref{sec:partial}.

\subsection{Variation combinations}

Some of the components described above can be combined to compose a more complex problem overall. Usually, one of these combinations will take components from different variation types, for example, simultaneous multiplicity of inputs and outputs. 

More specifically, there exist several studies involving MI ML scenarios \cite{miml,miml2}. In this case, examples from the input space are composed of several feature vectors and are associated to various labels. As a consequence, this model can represent many complicated problems where inputs and outputs have more structure than usual.

Other more uncommon situations are MV MI ML problems \cite{mvmiml}, where patterns have several instances which may or may not belong to the same space, a multi-output version of OR named graded ML classification \cite{graded-ml} and more complex input structures such as multi-layer MI MV \cite{mlmimv}, where a hierarchy of instances is present in each example. 

\section{Taxonomy}
\label{sec:taxonomy}

\begin{figure}[ht]
\centering
\includegraphics[width=.45\textwidth]{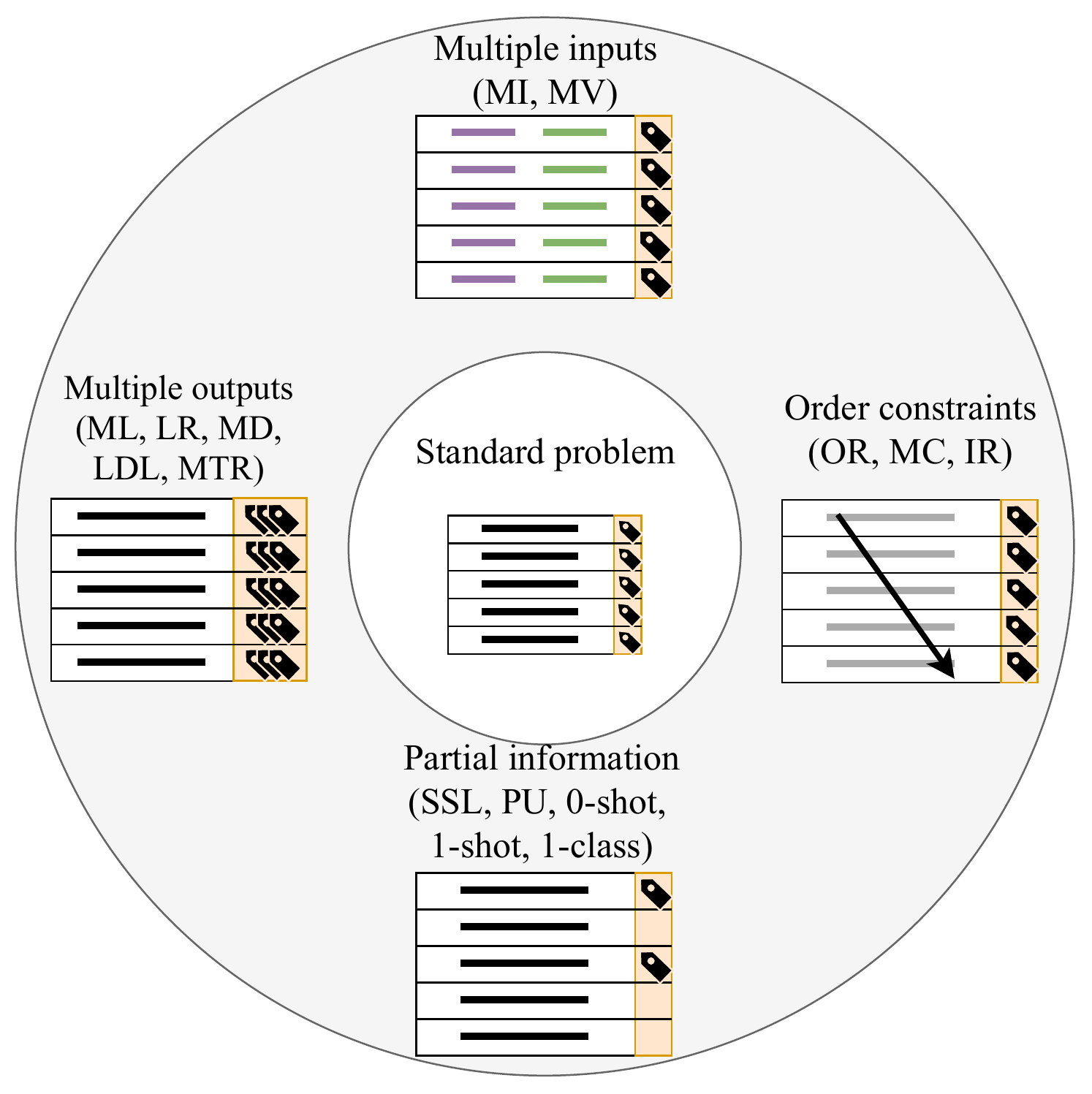}
\caption{\label{fig.multiples}Extensions of the standard supervised problem: multiple inputs or outputs, presence of orders and rankings, and partial information.}
\end{figure}

A first categorization of the variations analyzed in this work can be made according to how they differ from the standard problem. There can be multiplicity in the input space or the output space, order constraints may exist, or only partial information may be given in some cases. Fig.~\ref{fig.multiples} shows ways in which the traditional problems can be generalized. 

Problems introducing multiple inputs are MI and MV, whereas multiple outputs can be found on ML, MD, LR, LDL and MTR. Problems where orders are present are OR, MC and IR. Likewise, tasks with only partial information are, among others, semi-supervised learning, 
one-shot classification and zero-shot classification.

Finally, a generalized problem can be built out of combining several of these components: for example, a multiple-input multiple-output problem where the inputs and outputs can belong to structures like the ones defined above.

The rest of this section studies variations on the structure of the input space and output space, establishes relations among problems, and describes how they can be particularized or generalized to one another.

\subsection{Input structure}
\label{sec:multiinput}

In a standard supervised problem, the input space consists of single feature vectors and does not impose a specific order.

Problems where learning patterns are composed of multiple instances can usually be categorized into either MI, if the inputs share the same structure, or MV, otherwise. Their combination can also be considered as well, e.g. a problem where an example is composed of one or more photographs and one or more pieces of text. This would be a case of a MV MI problem.

There are also problems where there exists a partial or total order among instances, which is coupled with an order constraint in relation to the outputs. These are MC and IR.

\begin{figure}[ht]
\centering\scriptsize
\begin{tikzpicture}[node distance=.6cm, auto]
 \node[nobox] (title) {Input \mbox{structure} traits};
 \node[right=of title] (dummy1) {}
 edge[dummyhor] (title.east);
 \node[nobox, above=1cm of dummy1] (single) {Single \mbox{feature} vector}
 edge[] (dummy1.center);
 \node[right=of single] (dummy2) {}
 edge[dummyhor] (single.east);
 \node[nobox, above=of dummy2] (unord) {Unordered (standard)}
 edge[] (dummy2.center);
 \node[box, below=of dummy2] (ord) {Ordered (MC, IR)}
 edge[] (dummy2.center);
 \node[nobox, below=1.5cm of dummy1] (multiple) {Multiple \mbox{feature} vectors}
 edge[] (dummy1.center);
 \node[right=of multiple] (dummy3) {}
 edge[dummyhor] (multiple.east);
 \node[box, below=of dummy3] (same) {Same space (MI)}
 edge[] (dummy3.center);
 \node[box, above=of dummy3] (dif) {Different space (MV)}
 edge[] (dummy3.center);
\end{tikzpicture}
\caption{\label{fig.minputstr}Traits that can be found on the input structure of supervised problems.}
\end{figure}
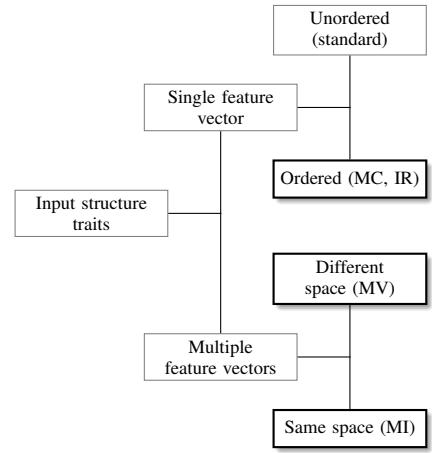

Fig.~\ref{fig.minputstr} summarizes these structural traits in a hierarchy and indicates problems where these traits are present.

\subsection{Output structure}
\label{sec:multioutput}

The diversity in output variations is higher than that of the input ones. A first sorting criterion is whether the codomain is discrete or continuous. This way, problems are either classification or regression ones.

Further subdivision of problems allows to separate these traits according to whether outputs remain scalars or become vectors. In the first case we consider order in the discrete scenario a nonstandard variation, which is present in OR and MC. In the second case, classification problems are spread into ML, LR and MD, and regression ones into LDL and MTR. 

\begin{figure*}[h]
\centering\scriptsize
\begin{tikzpicture}[node distance=.6cm, auto]
 \node[titlebox] (title) {Output structure traits};
 \node[dummy,below=of title] (dummy1) {}
 edge[dummyedge] (title.south);
 \node[nobox, left=2.75cm of dummy1] (discrete) {Discrete}
 edge[] (dummy1.center);
   \node[dummy,below=of discrete] (dummy2) {}
   edge[dummyedge] (discrete.south);
   \node[nobox, left=1.87cm of dummy2] (scalar1) {Scalar}
   edge[] (dummy2.center);
     \node[dummy,below=of scalar1] (dummy4) {};
     \node[nobox, left=.3cm of dummy4] (unord) {Unordered (standard classification)}
     edge[] (scalar1.south);
     \node[box, right=.3cm of dummy4] (ord) {Ordered (OR, MC)}
     edge[] (scalar1.south);
   \node[nobox, right=1.87cm of dummy2] (mult1) {Multiple}
   edge[] (dummy2.center);
     \node[box,below=of mult1] (dummy5) {Ranking (LR)}
     edge[] (mult1.south);
     \node[box, left=.3cm of dummy5] (unord) {Binary (ML)}
     edge[] (mult1.south);
     \node[box, right=.3cm of dummy5] (ord) {Finite (MD)}
     edge[] (mult1.south);
 \node[nobox, right=2.75cm of dummy1] (cont) {Continuous}
 edge[] (dummy1.center);
   \node[dummy,below=of cont] (dummy3) {}
   edge[dummyedge] (cont.south);
   \node[nobox, left=of dummy3] (scalar2) {Scalar (standard regression)}
   edge[] (dummy3.center);
   \node[nobox, right=of dummy3] (mult2) {Multiple}
   edge[] (dummy3.center);
     \node[dummy,below=of mult2] (dummy6) {};
     \node[box, left=.3cm of dummy6] (unord) {Distribution (LDL)}
     edge[] (mult2.south);
     \node[box, right=.3cm of dummy6] (ord) {Unrestricted (MTR)}
     edge[] (mult2.south);
\end{tikzpicture}
\caption{\label{fig.moutputstr}Traits that can be found on the output structure of supervised problems.}
\end{figure*}
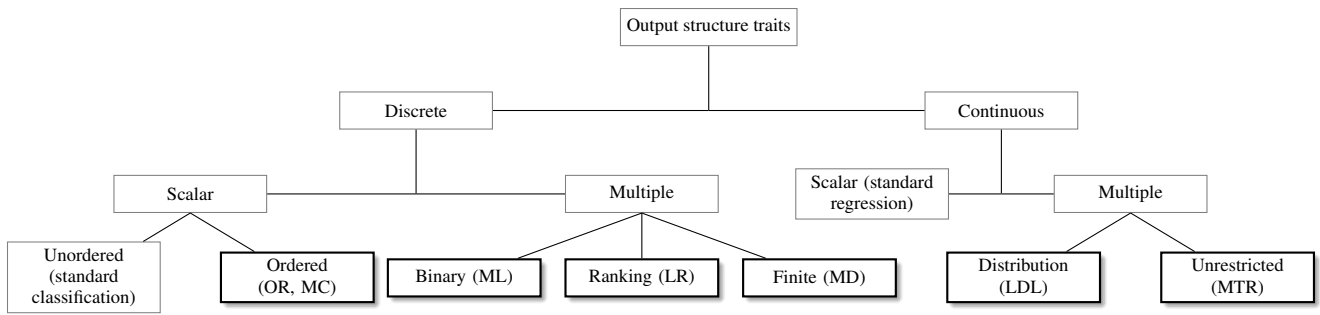

Fig.~\ref{fig.moutputstr} organizes these traits in a hierarchy based on the previous criteria. Each leaf of the tree also includes problems where each one is present.

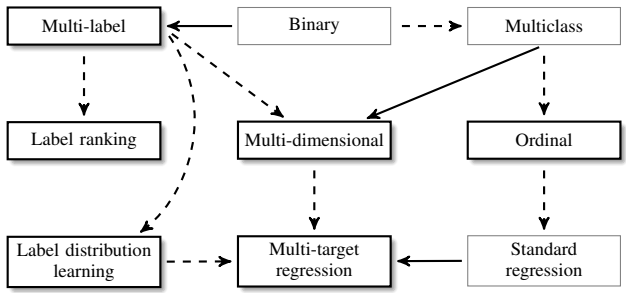
\begin{figure}[ht]
\centering\scriptsize
\begin{tikzpicture}[node distance=1cm, auto]
 
 \node[nobox] (bin) {Binary};
 \node[nobox, right=of bin] (multiclass) {Multiclass}
 edge[pil2] (bin.east);
 \node[box, below=of multiclass] (ord) {Ordinal}
 edge[pil2] (multiclass.south);
 \node[box, left=of bin] (mlabel) {Multi-label}
 edge[pil] (bin.west);
 \node[box, below=of mlabel] (lrank) {Label ranking}
 edge[pil2] (mlabel.south);
 \node[box, below=of bin] (mdim) {Multi-dimensional}
 edge[pil] (multiclass.south)
 edge[pil2] (mlabel.east);
 \node[nobox, below=of ord] (reg) {Standard regression}
 edge[pil2] (ord.south);
 \node[box, below=of mdim] (mtreg) {Multi-target regression}
 edge[pil2] (mdim.south)
 edge[pil] (reg.west);
 \node[box, below=of lrank] (ldl) {Label \mbox{distribution} learning}
 edge[pil2,->] (mtreg.west)
 edge[pil2,bend right=45] (mlabel.east);
\end{tikzpicture}
\caption{\label{fig.mouttax}Relations among supervised problems according to output structure. Arrows follow natural generalizations from one problem to another. Continuous arrows denote generalizations based on adding more variables of the same type. Dashed arrows indicate generalizations based on modifying existing target variables.}
\end{figure}

The variations in the structure of target spaces in supervised problems can be seen as generalizations of the standard problems. Furthermore, some of them are also more general than others. For example, ML problems can be seen as LR ones where, for a given instance, labels over a threshold are active and those below are not. Thus, LR is a generalization of the ML scenario. More relations of this kind are displayed in Fig.~\ref{fig.mouttax}.

As shown in the graph, an inclusion of more target variables of the same type transforms a binary problem into ML, a multiclass problem into MD and a single-target regression one into MTR. Similarly, inclusion of more values into each variable allows to generalize binary problems to multiclass, and ordinal to single-target regression, as well as ML ones to MD and these to MTR. LDL can be seen as a generalization of ML where real numbers between 0 and 1 are also allowed as values for a label. LR is a generalization of ML by the argument discussed before.

\subsection{Summary}
\label{sec:summary}

In this section input and output variations of standard supervised problems have been categorized and related. Table~\ref{tbl.identification} allows to identify specific problems according to which input and output traits are present.

\begin{table*} 
\centering\scriptsize
\setlength{\tabcolsep}{0.65em}
\begin{tabular}{r p{.16\textwidth} p{.20\textwidth} c p{.06\textwidth} p{.14\textwidth} p{.1\textwidth} p{.07\textwidth} }
\toprule
\multirow{ 3}{*}{\diaghead{Inputs~~~~~Outputs}{\scriptsize\textbf{Inputs}}{\scriptsize\textbf{Outputs}}} & \multicolumn{2}{c}{\textbf{Unordered outputs}}  &                  & \multicolumn{4}{c}{\textbf{Ordered outputs}}                    \\
& \multicolumn{1}{c}{\textbf{Scalar}} & \multicolumn{1}{c}{\textbf{Multiple}} &  & \multicolumn{2}{l}{\hphantom{Padding}\textbf{Scalar}} & \multicolumn{2}{c}{\textbf{Multiple}}  \\ 
&  & &  & \textbf{Discrete} & \textbf{Continuous} & \textbf{Discrete} & \textbf{Continuous} \\ 
\cmidrule{2-3}\cmidrule{5-8}
\vspace{.3em} \textbf{Unordered inputs}&
standard classification \cite{classification} & ML/MD classification \cite{mlc,mdc} & &
OR \cite{ord-survey} & standard regression \cite{regression} & Graded ML \cite{graded-ml} & MTR \cite{moutr} \\ \vspace{.3em}
\textbf{Ordered inputs}&
- &-  & &
MC \cite{mc-salva} & IR \cite{ir-book} &-  &-  \\ \vspace{.3em}
\textbf{Multiple instances}&
MI classification \cite{mil} & MIML/MIMD classification \cite{miml} & &- & MI regression \cite{mil} &- &- \\
\textbf{Multiple views}&
MV classification \cite{mviewl} & MVML/MVMD classification \cite{mvmiml} & &- & MV regression \cite{mviewl} &- &- \\
\bottomrule
\end{tabular}
\caption{\label{tbl.identification}Identification of problems according to their input traits (vertical axis) and output traits (horizontal axis).}
\end{table*}

\section{Common approaches to tackle nonstandard problems}
\label{sec:algorithms}

When tackling a nonstandard problem, most techniques follow one of two main approaches: problem transformation or algorithm adaptation. The first one relies on appropriate transformations of the data which result in one or more simpler, standard problems. The latter implies an extension or development of previously existing algorithms, in order to adapt them to the complexities induced by the structure of the data.

In the following subsections several methods based on both approaches are enumerated for each analysed problem.

\subsection{Problem transformation}

Problem transformation methods assume that a solution can be achieved by extracting one or more simpler problems out of the original one. For example, a problem with multi-dimensional targets could be transformed into many problems with scalar outputs. Then, these problems could be solved independently by a classical algorithm. A solution for the original problem would be the concatenation of those extracted from the simpler ones.

Next, the most common transformation techniques are described for each nonstandard supervised learning task previously introduced.

\paragraph{-- MI.} 
The taxonomy proposed in \cite{mic-taxonomy} describes an Embedded Space paradigm, where each bag is transformed into a single feature vector representing the relevant information about the whole bag. This transformation brings the MI problem into a single-instance one. Most of these methods are voca\-bulary-based, which means that the embedding uses a set of concepts to classify each bag according to its instances, resulting in a single vector with one component per concept.

\paragraph{-- MV.} Some naive transformations consist in ignoring every view except one, or concatenating feature vectors from all views, thus training a single-view model in both cases \cite{mv-spectral}. A preprocessing based on Canonical Correlation Analysis \cite{mv-cca} is able to project data from multiple views onto a lower-dimensional, single-view space.

\paragraph{-- ML.} 
Transformation methods for ML classification \cite{mlmethods} are diverse: Binary Relevance trains separate binary classifiers for each label. Label Powerset reduces the problem to a multiclass one by treating each individual labelset as an independent class label, and Random k-Labelsets \cite{ml-rakel} extracts an ensemble of multiclass problems similarly. Classifier chains \cite{ml-chains} trains subsequent binary classifiers accumulating previous predictions as inputs. ML problems can also be transformed to LR \cite{ml-clr}.

\paragraph{-- MD.} In some cases, independent classifiers can be trained for several dimensions \cite{mdc,mdc-indep} but this method ignores possible correlations among dimensions. An alternative transformation, building a different label from each combination of classes, would produce a much larger label space and thus is not typically applied.

\paragraph{-- LDL.} A LDL problem can be reduced to multiclass classification by extracting as many single-label examples as labels for each one of the training instances \cite{ldl}. These new examples are assigned a class corresponding to each label and weighted according to its degree of description. During the prediction process, the classifier must be able to output the score/confidence for each label, which can be used as its description degree.

\paragraph{-- LR.} A reduction of this problem to several binary problems can be achieved by learning pairwise preferences \cite{lrankpairwise}. This transforms a $c$-label problem into $c(c-1)/2$ binary problems describing a comparison among two labels. An alternative reduction by means of constraint classification \cite{lr-constraint} builds a single binary classification dataset by expanding each label preference into a new positive instance and a new negative instance. The feature space of the new binary problem has dimension $nc$, where $n$ is the original dimension and $c$ the number of labels, due to the constraints embedded in it by Kesler's construction \cite{nilsson}.

\paragraph{-- MTR.} There are several ways to transform a MTR problem into several single-target regression ones. Some of them are inspired by the ML field, such as a one-vs-all single-target reduction, multi-target stacking and regressor chains \cite{mtrviaml}. All of them train single-target regressors for several extracted problems, and then combine the obtained predictions. A different approach based on support vectors \cite{mtr-lssvr} extends the feature space which expresses the multi-output problem as a single-target one that can be solved using least squares support vector regression machines.

\paragraph{-- OR.} An ordinal problem with $c$ classes can be transformed into $c-1$ binary classification problems by using each class from the second to the last one as a threshold for the positive class \cite{ord-simple}. This decomposition can be called \emph{ordered partitions} and is not the only possible one: others are \emph{one-vs-next}, \emph{one-vs-followers} and \emph{one-vs-previous} \cite{ord-survey}. Several 3-class problems can also be obtained by using, for the $i$-th problem, classes ``$l_i$'', ``$<l_i$'' and ``$>l_i$''.

\paragraph{-- MC.} The authors in \cite{monotonicity} describe a procedure to tackle binary MC problems by means of IR. Multiclass MC cases can be reduced to several binary MC ones, which in turn are solved as IR problems. 

\subsection{Algorithm adaptation}

Existing methods for classical problems can be extended in order to introduce the necessary complexities of nonstandard variations. As an example, nearest neighbor methods could be coupled with new distance metrics in order to be able to measure similarity among multiple inputs.

The rest of this section presents some algorithm adaptations which can be used to tackle nonstandard supervised tasks.

\paragraph{-- MI.} 
Methods that work on instance level are adaptations of algorithms from single-instance classification whose responses are then aggregated to build the bag-level classification \cite{mic-taxonomy}. They typically assume that one positive instance implies a positive bag. Adaptations of common algorithms have been proposed with support vector machines (SVM) \cite{mi-svm} and neural networks \cite{mi-nn}, whereas some original methods in this area are Axis-Parallel Rectangles \cite{mi-apr} and Diverse Density \cite{mi-framework}. In the bag-space paradigm, methods treat bags as a whole and use specific distance metrics with distance as well as kernel-based classifiers, such as k-nearest neighbor (k\nobreakdash-NN) \cite{mi-knn} or SVM \cite{mi-kernel}.

\paragraph{-- MV.} Supervised methods for MV are comparatively less developed than semi-supervised ones. Nonetheless, there is an extension of SVM \cite{mv-svm} which simultaneously looks for two SVMs, one in each of the feature spaces of a two-view problem. There is an extension of Fisher discriminant analysis as well \cite{mv-fda}.

\paragraph{-- ML.} The most relevant algorithm adaptations \cite{mlmethods} are based on standard classification algorithms with added support for choosing more than one class at a time: adaptations exist for k-NN \cite{ml-knn}, decision trees \cite{ml-dt}, SVMs \cite{ml-svm}, association rules \cite{ml-rules} and ensembles \cite{mlensembles}.

\paragraph{-- MD.} Specific Bayesian networks have been proposed for the MD scenario \cite{md-bayes,md-bayes2}, as well as Maximum Entropy-based algorithms \cite{mdc,mdc-indep}.

\paragraph{-- LDL.} Proposals in \cite{ldl} are adaptations of k-NN, with a special derivation of the label distribution of an unseen instance given its neighbors, and backpropagated neural networks, where the output layer indicates the label distribution of an instance. Other proposed methods are based on the optimization algorithms BFGS and Improved Iterative Scaling. 

\paragraph{-- LR.} Boosting methods have been adapted to LR \cite{lr-boost}, as well as the SVM proposed in \cite{ml-svm} for ML which can be naturally extended to LR \cite{lranksurvey}. An adaptation of online learning algorithms such as the perceptron has also been developed \cite{lr-online}.

\paragraph{-- MTR.} First methods able to treat MTR problems were actually generalizations of statistical methods for single-target regression \cite{mtr-rank,mtr-canon}. Other common methods which have been extended to predict multiple regression variables are support vector regression \cite{mtr-svr1,mtr-svr2}, kernel-based methods \cite{mtr-kern1,mtr-kern2}, and regression trees \cite{mtr-trees} as well as random forests \cite{mtr-rf}.

\paragraph{-- OR.} Neural networks can be used to tackle OR with slight changes in the loss function or the output layer \cite{or-nn,or-nn2}. Similarly, extreme learning machines have also been applied to this problem \cite{or-elm,or-elm2}. Common techniques such as k-NN or decision trees have been coupled with global constraints for OR \cite{or-knn-dt}, and extensions of other well known algorithms such as Gaussian processes \cite{or-gp} and AdaBoost \cite{or-ada} have been proposed as well.

\paragraph{-- MC.} Algorithm adaptations generally take a well known technique and add monotonicity constraints. For example, there exist in the literature adaptations of k-NN \cite{mc-knn}, decision trees \cite{mc-trees}, decision rules \cite{mc-rules,mc-rules2} and artificial neural networks \cite{mc-monnets}.

\vspace{1em}

Table~\ref{tbl.methods} gathers all the methods described previously to tackle nonstandard supervised tasks.

\begin{table}[ht]
\centering\scriptsize
\setlength{\tabcolsep}{0.55em}
\renewcommand{\arraystretch}{1.4}
\begin{tabular}{r p{.2\textwidth} p{.2\textwidth}}
\toprule
\textbf{Task} & \textbf{Problem transformation} & \textbf{Algorithm adaptation} \\ \cmidrule{1-3}
\textbf{MI} & 
Embedded-space \cite{mic-taxonomy} & 
SVM \cite{mi-svm,mi-kernel}\newline Neural networks \cite{mi-nn}\newline k-NN \cite{mi-knn} \\
\textbf{MV} & 
Canonical correlation analysis \cite{mv-cca}& 
SVM \cite{mv-svm} \newline Fisher discriminant analysis \cite{mv-fda}\\
\textbf{ML} & 
Binary Relevance \cite{mlmethods} \newline Label Powerset \cite{mlmethods} \newline Classifier chains \cite{ml-chains} & 
k-NN \cite{ml-knn} \newline Decision trees \cite{ml-dt}\newline SVM \cite{ml-svm} \newline Association rules \cite{ml-rules} \newline Ensembles \cite{mlensembles}\\
\textbf{MD} & 
Independent classifiers \cite{mdc,mdc-indep} & 
Bayesian networks \cite{md-bayes,md-bayes2}\newline Maximum Entropy \cite{mdc,mdc-indep} \\
\textbf{LDL} & 
Multiclass reduction \cite{ldl} & 
k-NN \cite{ldl} \newline Neural networks \cite{ldl}\\
\textbf{LR} & 
Pairwise preferences \cite{lrankpairwise} \newline Constraint classification \cite{lr-constraint} &
Boosting \cite{lr-boost} \newline SVM \cite{lranksurvey} \newline Perceptron \cite{lr-online}\\
\textbf{MTR} &
               ML 
               \cite{mtrviaml} \newline Support vectors \cite{mtr-lssvr} & 
Generalizations \cite{mtr-rank,mtr-canon} \newline Support vector regression \cite{mtr-svr1,mtr-svr2}\newline Kernel-based \cite{mtr-kern1,mtr-kern2} \newline Regression trees \cite{mtr-trees} \newline Random forests \cite{mtr-rf}\\
\textbf{OR} & 
Ordered partitions \cite{ord-simple} \newline One-vs-next, One-vs-followers, One-vs-previous \cite{ord-survey} \newline 3-class problems \cite{ord-survey}& 
Neural networks \cite{or-nn,or-nn2} \newline Extreme learning machines \cite{or-elm,or-elm2} \newline Decision trees \cite{or-knn-dt} \newline Gaussian processes \cite{or-gp} \newline AdaBoost \cite{or-ada} \\
\textbf{MC} & 
Reduction to IR \cite{monotonicity}& 
k-NN \cite{mc-knn}\newline Decision trees \cite{mc-trees}\newline Decision rules \cite{mc-rules,mc-rules2}\newline Neural networks \cite{mc-monnets}\\
\toprule
\end{tabular}
\caption{\label{tbl.methods}Summary table of presented methods according to their type of approach.}
\end{table}

\section{Applications. Original real word scenarios}
\label{sec:applications}

The problems studied in this work have their origins in real-world scenarios which are related below:

\paragraph{-- MI.} Problems modeled under MI learning are drug activity prediction \cite{mi-apr}, where each pattern describes a molecule and its different forms are represented by instances; image classification \cite{mic-taxonomy}, and bankruptcy \cite{ami-bank}. Most of the datasets used in experimentations, however, are usually synthetic.

\paragraph{-- MV.} Some situations where data is described in multiple views are multilingual text categorization \cite{amv-multilingual}, face detection with several poses \cite{amv-face}, user localization in a WiFi network \cite{amv-wifi}, advertisements described by their image and surrounding text \cite{amv-ads-webkb} and image classification with several color-based views and texture-based views \cite{amv-corel}.

\paragraph{-- ML.} Problems which fall naturally under the ML definition are text classification under several categories simultaneously \cite{aml-text}, image labeling \cite{aml-scene}, question tagging in forums where tags can co-exist \cite{aml-question}, protein classification \cite{aml-protein}.

\paragraph{-- MD.} Applications of MD classification include classification of biomedical text \cite{mdc}, where predicted dimensions for a given document are its focus, evidence type, certainty level, polarity and trend; gene function identification \cite{md-bayes}; tumor classification, and illness diagnosis in animals \cite{md-bayes2}.

\paragraph{-- LR.} The field known as \emph{preference learning} has been gaining interest \cite{lrankpairwise}, and LR is one of the problem that falls under this term. LR is also frequently applied in ML scenarios \cite{lrank4ml}, where a threshold can be applied in order to transform an obtained ranking into a labelset.

\paragraph{-- LDL.} Data with relative importance of each label appears in applications such as analysis of gene expression levels in yeast \cite{aldl-yeast}, or emotion description from facial expressions \cite{aldl-face}, where a face can depict several emotions in different grades.

\paragraph{-- MTR.} Applications modeled as MTR problems are diverse, including modeling of vegetation condition in ecosystems assigning several scores which depend on the vegetation type \cite{amtr-eco}, prediction of audio spectrums of wind tunnel tests \cite{amtr-wind}, and estimation of several biophysical parameters from remote sensing images \cite{amtr-remote}.

\paragraph{-- OR.} The most salient fields where OR can be found are text classification \cite{aor-text}, where the predicted variable may be an opinion scale or a degree of satisfaction; image categorization \cite{aor-image}; medical research \cite{aor-medical}; credit rating \cite{aor-credit}, and age estimation \cite{aor-age}.

\paragraph{-- MC.} Monotonicity constraints are found in problems related to customer satisfaction analysis \cite{amc-customer}, in which overall appreciation of a product must increase along with the evaluation of its features; house pricing \cite{mc-trees}; bankruptcy risk evaluation \cite{amc-bank}, and cancer prediction \cite{amc-cancer}, among others.

\section{Other nonstandard variations}
\label{sec:othervariations}

This section covers variations of the standard supervised problem which are further from the central focus of this paper less related to those above. 

\subsection{Learning with partial information}
\label{sec:partial}

In a standard supervised classification setting, it is assumed that every training example is labeled accordingly and that there exist examples for every class that may appear in the testing phase. When only a fraction of the training instances are labeled, the problem is considered semi-supervised \cite{semi-sup}, but generally there still exist labeled samples for each class.

In positive-unlabeled learning \cite{pu-learn,pu-text}, however, labeled examples provided within the training set are only positive. This means the learning algorithm only knows about the class of positive instances, and unlabeled ones can have either class. 

A different scenario arises when the training set only consists of negative (or only positive) instances, and no unlabeled examples are provided. This is known as one-class classification \cite{oneclass}, and data of this nature can be obtained from outlier detection applications, where positive examples are hardly recorded.

A problem which may be seen as a generalization of one-class classification is zero-shot learning \cite{zeroshot}, a situation where unseen classes are to be predicted in the testing stage. That is, the label space $Y$ includes some values which are not present in any training pattern, but the classifier must be able to predict them. For example, if in a speech recognition problem $Y$ is the set of all words in English, the training set is unlikely to have at least one instance for each word, thus the classifier will only succeed if it is capable of assigning unlearned words to test examples.

A relaxation on the obstacles of zero-shot learning is present in one-shot learning \cite{oneshot}, where algorithms attempt to generalize from very few (1 to 5) examples of each class. This is a common circumstance in the field of image classification, where the cost of collecting and labeling data samples is high.

A classification of these problems according to the type of missing information can be found in Table~\ref{tbl.partial-problems}.

\begin{table}[ht]
\centering
\renewcommand{\arraystretch}{1.4}
\begin{tabular}{r  p{0.16\textwidth}}
\toprule
\textbf{Trait} & \textbf{Problem types} \\ \hline
Presence of unlabeled instances & Semi-supervised \cite{semi-sup} \newline Positive-unlabeled \cite{pu-learn} \\
No representation of some classes &   One-class \cite{oneclass} \newline Positive-unlabeled \cite{pu-learn} \newline Zero-shot \cite{zeroshot} \\
Scarce representation of some classes &  One-shot \cite{oneshot}\\
\toprule
\end{tabular}
\caption{\label{tbl.partial-problems}Partial information problems according to the kind of absence in the training set.}
\end{table}

\subsection{Prediction of structured data}

The nonstandard variations described in this work generalize traditional supervised problems where the predicted output is at most a vector whose components take values in either a finite set or $\mathbb R$. Further generalizations are possible if other kinds of structures are allowed. For example, the target may take the form of an ordered sequence or a tree. In this case, the problem usually enters the scope of structured prediction \cite{str-pred}, a generalization of supervised learning where methods must build structured data associated to input instances.

A particular case of supervised problem which can be seen under the umbrella of structured prediction is learning to rank \cite{ltr}, which does not involve a label space as such. Instead, training consists in learning from a set of feature vectors with a series of preferences among them, that is, a partial or total order in the training set. During testing a set of feature vectors is provided and the desired output is a ranking (with a predefined number of relevance levels, allowing ties) or a sorting (simply an ordering of the instances). This problem differs from OR in that individual classifications are usually meaningless: only relative distances among ranked instances matter.

\section{Conclusions}
\label{sec:conclusions}

Traditional supervised learning comprises two well known problems in machine learning: classification and regression. However, the multitude of applications which do not strictly fit the structure of the standard versions of those problems have favored the development of alternative versions which are more flexible and allow the analysis of more complex situations. 

In this work an overview of nonstandard variations of supervised learning problems has been presented. A novel taxonomy under several criteria has described relationships among these variations, where the main differentiating properties are multiplicity of inputs, multiplicity of outputs, presence of order relations and constraints, and partial information. Afterwards, common methods for tackling these problems have been outlined and their main applications have been mentioned as well. Finally, some additional variants which were left out of the scope of the previous analysis have been introduced as well.

Design of novel algorithms for nonstandard supervised tasks is scarcer than adaptations and transformations, but there exist some approximations and even more open possibilities for tackling these from classical algorithmic perspectives, such as probabilistic and heuristic methods, information theory and linear algebra, among others. 



\begin{acknowledgements}
  D. Charte is supported by the Spanish Ministry of Science, Innovation and Universities under the FPU National Program (Ref. FPU17/04069). This work has been partially supported by projects TIN2017-89517-P (FEDER Founds) of the Spanish Ministry of Economy and Competitiveness and TIN2015-68454-R of the Spanish Ministry of Science, Innovation and Universities.
\end{acknowledgements}



\end{document}